\documentclass{article}

     \usepackage[final,nonatbib]{tackling_climate_workshop_style}

\usepackage[utf8]{inputenc} % allow utf-8 input
\usepackage[T1]{fontenc}    % use 8-bit T1 fonts
\usepackage{hyperref}       % hyperlinks
\usepackage{url}            % simple URL typesetting
\usepackage{booktabs}       % professional-quality tables
\usepackage{amsfonts}       % blackboard math symbols
\usepackage{nicefrac}       % compact symbols for 1/2, etc.
\usepackage{microtype}      % microtypography

\usepackage{amsmath}

\usepackage{cite}
\usepackage{subcaption}
\usepackage{cleveref}
\usepackage{graphicx}
\usepackage{tabularx}

\usepackage{dirtytalk}
\usepackage{xcolor}

\definecolor{blue_source}{rgb}{0.4, 0.55, 0.7}
\definecolor{orange_sink}{rgb}{0.925,0.447,0.2}

\usepackage{tikz}
\usetikzlibrary{calc,matrix,positioning,arrows,decorations.pathreplacing,angles,quotes,math}

\title{Towards dynamic stability analysis of sustainable power grids using graph neural networks}

\author{%
  Christian Nauck \\
  Department 4 Complexity Sciences \\
  Potsdam Institute for Climate Impact Research\\
  \texttt{nauck@pik-potsdam.de} \\
  \And
  Michael Lindner \\
  Department 4 - Complexity Sciences \\
  Potsdam Institute for Climate Impact Research\\
  \texttt{mlindner@pik-potsdam.de} \\
  \And
  Konstantin Schürholt \\
  AIML Lab, School of Computer Science\\
  University of St.Gallen\\
  \texttt{konstantin.schuerholt@unisg.ch} \\
  \And
    Frank Hellmann \\
  Department 4 Complexity Sciences \\
  Potsdam Institute for Climate Impact Research\\
  \texttt{hellmann@pik-potsdam.de} \\
}

\begin{document}

\maketitle

\begin{abstract}
To mitigate climate change, the share of renewable needs to be increased. Renewable energies introduce new challenges to power grids due to decentralization, reduced inertia and volatility in production. The operation of sustainable power grids with a high penetration of renewable energies requires new methods to analyze the dynamic stability. We provide new datasets of dynamic stability of synthetic power grids and find that graph neural networks (GNNs) are surprisingly effective at predicting the highly non-linear target from topological information only. To illustrate the potential to scale to real-sized power grids, we demonstrate the successful prediction on a Texan power grid model.

\end{abstract}

\section{Introduction}

%%%NEW
Adaption to and mitigation of climate change jointly influence the future of power grids: 
1) Mitigation of climate change requires power grids to be carbon-neutral, with the bulk of power supplied by solar and wind generators. These are more decentralized and have less inertia than traditional power generators and their production is more volatile. Hence, sustainable power grids operate in different states and the frequency dynamics need to explored in more detail.
2) A higher global mean temperature increases the likelihood as well as the intensity of extreme weather events such as hurricanes or heatwaves \cite{field_managing_2012,portner_ipcc_2022} which result in great challenges to power grids. 
%Such events may disturb the operation of the power grid in a myriad of hard to predict ways, e.g. when a heatwave leads to a surge in energy consumption due to air conditioning or a storm destroying power lines.
Building a sustainable grid as well as increasing the resilience of existing power grids towards novel threats are challenging tasks on their own. Tackling climate change in the power grid sector calls for a solution to both at the same time and requires for new methods to investigate the dynamic stability. \looseness-1
%%%

%   mitigation: renewable energies, adaptation: resilience to extreme weather events -> pioneering work, simplifications, method development
  
%% The following paragraphs justify our way of power grid modeling. We could instead offload that work to the literature and just motivate high computational effort as the concrete challenge that we tackle with GNNs?
  
% more renewable energies -> challenges
%Increasing the share of renewable energies in total energy production is one of the key targets on the path to carbon-neutral societies. In contrast to conventional power plants, renewable energies are more decentralized, have less inertia and their production is more volatile. Those aspects pose challenges to the current power grid infrastructure, both in terms of grid expansion and stable operation with large shares of renewable energies. Renewable energies will have to start contributing to the dynamical stability of the system \cite{milano_foundations_2018,christensen_high_2020} in the future, requiring a new understanding of complex dynamics.

% This paper is concerned with the stable operation of future power grids.

Predicting the dynamic stability is a challenging task and grid operators are currently limited to analyze individual contingencies in the current state of the transmission grid only. %, because 
Conducting high-fidelity simulations of the whole dynamic hierarchy of the power grid and exploring all possible states is not feasible \cite{liemann_probabilistic_2021}. For future power grids an understanding of how to design robust dynamics is required. This has led to a renewed interdisciplinary interest in understanding the collective dynamics of power grids \cite{brummitt_transdisciplinary_2013}, with a particular focus on the robustness of the self-organized synchronization mechanism underpinning the stable power flow \cite{rohden_self-organized_2012,motter_spontaneous_2013,dorfler_synchronization_2013} by physicists and control mathematicians\cite{witthaut_collective_2022}.\looseness-1

To understand which structural features impact the self-organized synchronization mechanism, it has proven fruitful to take a probabilistic view. Probabilistic approaches are well established in the context of static power flow analysis \cite{borkowska_probabilistic_1974}. In the dynamic context, considering the expected likelihood of failure given a non-linear, random perturbation effectively averages over the various contingencies. Such probabilities are thus well suited to reveal structural features that enhance the system robustness or vulnerability. This approach has been highly successful in identifying particularly vulnerable grid regions \cite{menck_how_2014, schultz_detours_2014, nitzbon_deciphering_2017} and revealing general mechanisms of desynchronization \cite{hellmann_network-induced_2020}. Probabilistic stability assessments recently got more attention in the engineering community as well \cite{liu_quantifying_2017,liu_basin_2019,liemann_probabilistic_2021}.\looseness-1

Assessing probabilistic dynamic stability of a given class of power grid models is computationally expensive. Further, the probabilistic dynamics are a sensitive function of the structural variables and minor changes like the addition of a single power line may lead to very different outcomes (see e.g. \cite{witthaut_braesss_2012}). Since the space of parameters that may influence the dynamic stability of a power grid is very large, an explicit computational assessment of all potential configurations is impossible. If graph neural networks (GNNs) were able to reliably predict probabilistic dynamic stability, they could be used to select promising candidate configurations for which a more detailed assessment should be carried out. Moreover, the analysis of the decision process of ML-models might lead to new unknown relations between dynamical properties and structural aspects, such as the grids topology or the distribution of loads and generators. Such insights may ultimately inform the design and development of power grids.\looseness-1

\begin{figure}[t!]
    \centering
    \vspace{-1pt}
    \includegraphics[width=\linewidth]{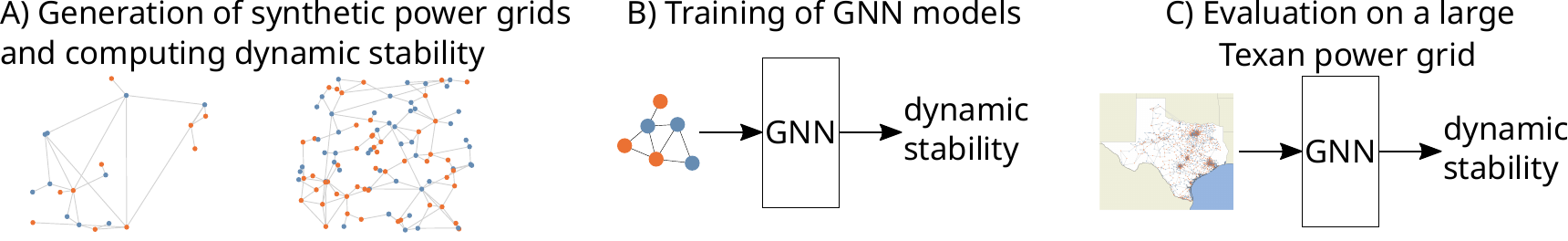}
    % \caption{This papers' approach towards predicting dynamic stability  of power grids.}
    \caption{Our approach towards predicting dynamic stability  of power grids by using synthetically generated grids, training GNNs and evaluating the methods on a real-sized power grid.}
    \vspace{-11pt}
	\label{fig_scheme}
\end{figure}

% \paragraph{Related work on power grid property prediction}
\paragraph{Related work:}
Since power grids have an underlying graph structure, the recent development of graph representation learning \cite{bronstein_geometric_2021, hamilton_graph_2020} introduces promising methods to use machine learning in the context of power grids. There is a number of applications using GNNs for different power flow-related tasks \cite{donon_graph_2019,kim_graph_2019,bolz_power_2019,retiere_spectral_2020,wang_probabilistic_2020,owerko_optimal_2020,gama_graph_2020,misyris_physics-informed_2020,liu_searching_2021,bush_topological_2021,liu_guiding_2020,jhun_prediction_2022} and to predict transient dynamics in microgrids \cite{yu_pidgeun_2022}. 
%%%NEW
%In their pioneering work Nauck et al.~\cite{nauck_predicting_2022} explore the potential of GNNs to aid the energy transition
In \cite{nauck_predicting_2022} the potential of GNNs is explored to aid the energy transition by reducing the computational effort of dynamical stability assessment for future power grids. We expand this work by training larger GNN models on more data and by evaluating them on a synthetic model of the Texan power grid. Our new GNN models outperform the benchmark models defined in~\cite{nauck_predicting_2022} as well as established ML methods like linear regression and multilayer perceptrons (MLPs).\looseness-1
\paragraph{Contributions}
We introduce new datasets of probabilistic dynamic stability of synthetic power grids. The new datasets have 10 times the size of previously published ones and and include a Texan power grid model to map the path towards real-world applications. We evaluate the predictive performance of GNNs and systematically compare those against baselines.
Our results demonstrate i) that the larger dataset allows to train more powerful GNNs, (ii) which outperform the baselines, and (iii) transfer from the new datasets to a real-sized power grid. The general approach of this paper is visualized in \Cref{fig_scheme}. \looseness-1

%We introduce new datasets of probabilistic dynamic stability of synthetic power grids as new challenges for GNN models. The new datasets have 10 times the size of previously published ones and and include a Texan power grid model to map the path towards real-world applications. We train benchmark models to evaluate the difficulty of the task. We demonstrate that the larger dataset size allows to train larger models and achieve higher performance. We successfully apply the methods to a real-world-sized power grid with a different distribution of target values. The general approach is visualized in \Cref{fig_scheme}.

% \begin{figure}[h]

\section{Generation of the datasets}
\label{sec_generation_dataset}

\paragraph{Power grids as dynamical systems}
From a dynamical systems point of view, the central challenge addressed in this work is characterizing the interplay of network topology with the dynamics of synchronizing oscillators. The dynamics of real-world power grids feature synchronizing oscillators coupled on complex networks. However, the full dynamics certainly contain many more aspects. A full scale analysis that can treat high-fidelity models of real systems is currently out of reach for several reasons. These include that real world data does not exist or is not accessible, synthetically generating large numbers of realistic dynamical models is challenging, and that models can not be simulated fast enough with current software \cite{liemann_probabilistic_2021}. These problems force trade-offs on us, most notably reducing the details of the intrinsic behaviors of oscillators to the inertial Kuramoto model~\cite{kuramoto_self-entrainment_1975, bergen_structure_1981}. As a consequence, our results are not directly transferable to real-world applications. Nevertheless, any future treatment that moves towards more accurate dynamical models will also have to solve the challenging subproblem of the impact of topology on synchrony that we consider here.\looseness-1

\paragraph{Dynamic stability of power grids}
We quantify probabilistic dynamic stability with the widely used non-linear measure single-node basin stability (SNBS) \cite{menck_how_2013}. SNBS measures the probability that the entire grid asymptotically returns to a stable state after initial perturbations at single nodes. Crucially, it is not the result of a single simulation but describes a statistical behaviour (expected value of a Bernoulli experiment). SNBS takes values in $[0,1]$, with higher values indicating more stability. \looseness-1
%If all applied perturbations return to a stable operational state, SNBS would be 1. 
%Even though the perturbations are applied at a single node, this measure captures the reaction of the entire grid and hence is not a purely local property. 
%To sum up, the probabilistic measure SNBS captures non-linear, dynamic effects after nodal perturbations.

\begin{figure}
    \centering
    \vspace{-4pt}
        \includegraphics[width=.8\linewidth]{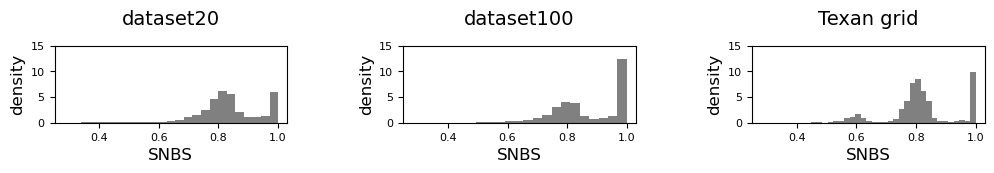}
    \caption{Normalized distributions of the target values (SNBS) in our datasets.}
    \vspace{-8pt}
	\label{fig_gridExamples_distributions}
\end{figure}

\paragraph{Procedure to generate the datasets}
%For the generation of the datasets, w
We closely follow the work in \cite{nauck_predicting_2022} and extend this by computing 10 times as many grids. To investigate different topological properties of differently sized grids, we generate two datasets with either 20 or 100 nodes per grid, referred to as dataset20 and dataset100. To enable the training of complex models, both datasets consist of 10,000 graphs. The simulations take more than 550,000 CPU hours. Additionally, we provide the dynamic stability of a real-sized  model of the Texan power grid, consisting of $1,910$ nodes, to take a first step towards real-world applications. We use the synthetic model by Birchfield et al.~\cite{birchfield_grid_2017} since real power grid data is not available due to security reasons. Even with our simplified modeling the simulation of that large grid takes 127,000 CPU hours, highlighting the potential %benefits of fast and reliable predictions with GNNs. 
of fast predictions with GNNs.
\paragraph{Properties of the datasets}
\label{sec_design_choices}
Examples of the grids of dataset20, dataset100, and the Texan power grid are given in \Cref{fig_scheme} (A). The distributions of SNBS characterized by multiple modes are given in \Cref{fig_gridExamples_distributions}. Interestingly, the SNBS distribution of the Texan power grid has a third mode which is challenging for prediction tasks.
Overall, the power grid datasets consist of the adjacency matrix and the binary injected power $P$ per node as inputs, and nodal SNBS as target values.

\section{Experimental setup to predict SNBS of power grids using GNNs}
% \subsection{Experimental setup}
On both datasets, we train GNNs and baselines on nodal regression tasks. The power grids are represented by the adjacency matrix and a binary feature vector representing sources and sinks. Both are fed into GNNs as input. GNNs are trained to predict SNBS for each node (\Cref{fig_scheme} B). We split the datasets in training, validation and test sets (70:15:15). The validation set is used for the hyperparameter optimization, we report the performance on the test set. To minimize the effect of initializations we use 5 different initializations per model and consider the three best to compute average performances.\looseness-1

We analyze the performance of different GNN models: ArmaNet \cite{bianchi_graph_2021}, GCNNet \cite{kipf_semi-supervised_2017}, SAGENet \cite{hamilton_inductive_2017} and TAGNet \cite{du_topology_2017}. The details of the models based on a hyperparameter study are in \Cref{sec_hyperparameter}.
%based on several GNN-architectures based on the following types of convolution: ARMA filters by \cite{bianchi_graph_2021}, Graph Convolutional Networks (GCN) by \cite{kipf_semi-supervised_2017}, SAmple and aggreGatE (SAGE) by \cite{hamilton_inductive_2017} and Topology Adaptive Graph Convolution (TAG) by \cite{du_topology_2017}. We refer to the models by ArmaNet, GCNNet, SAGENet and TAGNet.
%We conduct hyperparameter studies to optimize the model structure, see \Cref{sec_hyperparameter} for details.
To evaluate the performance, we use the coefficient of determination ($R^{2}$-score, see \Cref{app:r2}).\looseness-1
% To evaluate the performance, we use the coefficient of determination ($R^{2}$-score, see \Cref{app:r2}).

%We analyze the performance of different models based on several GNN-architectures based on the following types of convolution: ARMA filters by \cite{bianchi_graph_2021}, Graph Convolutional Networks (GCN) by \cite{kipf_semi-supervised_2017}, SAmple and aggreGatE (SAGE) by \cite{hamilton_inductive_2017} and Topology Adaptive Graph Convolution (TAG) by \cite{du_topology_2017}. We refer to the models by ArmaNet, GCNNet, SAGENet and TAGNet. We conduct hyperparameter studies to optimize the model structure, see \Cref{sec_hyperparameter} for details. To evaluate the performance, we use the coefficient of determination ($R^{2}$-score, see \Cref{app:r2}).

As baselines we use linear regression and two differently sized MLPs: MLP1 and MLP2. The inputs for MLP and the regression are based on network measures according to Schultz et al.~\cite{schultz_detours_2014}. Details regarding the MLPs and network measures are in \Cref{sec_regresssion_MLP}. Furthermore, we use the best GNN from \cite{nauck_predicting_2022} as additional baseline and call this model Ar-bench.

\tikzstyle{output}=[very thick, minimum size=1.5em, draw=white!100, fill=white!100, minimum width=2.5em, minimum height = 5em]
\tikzstyle{layer}=[very thick, minimum size=1.5em, draw=black!100, fill=white!100, minimum width=2.5em, minimum height = 6em]

\section{Results of predicting dynamic stability}
\begin{figure}[h]
    \centering
    \begin{subfigure}{.49\linewidth}
        \includegraphics[width=\linewidth]{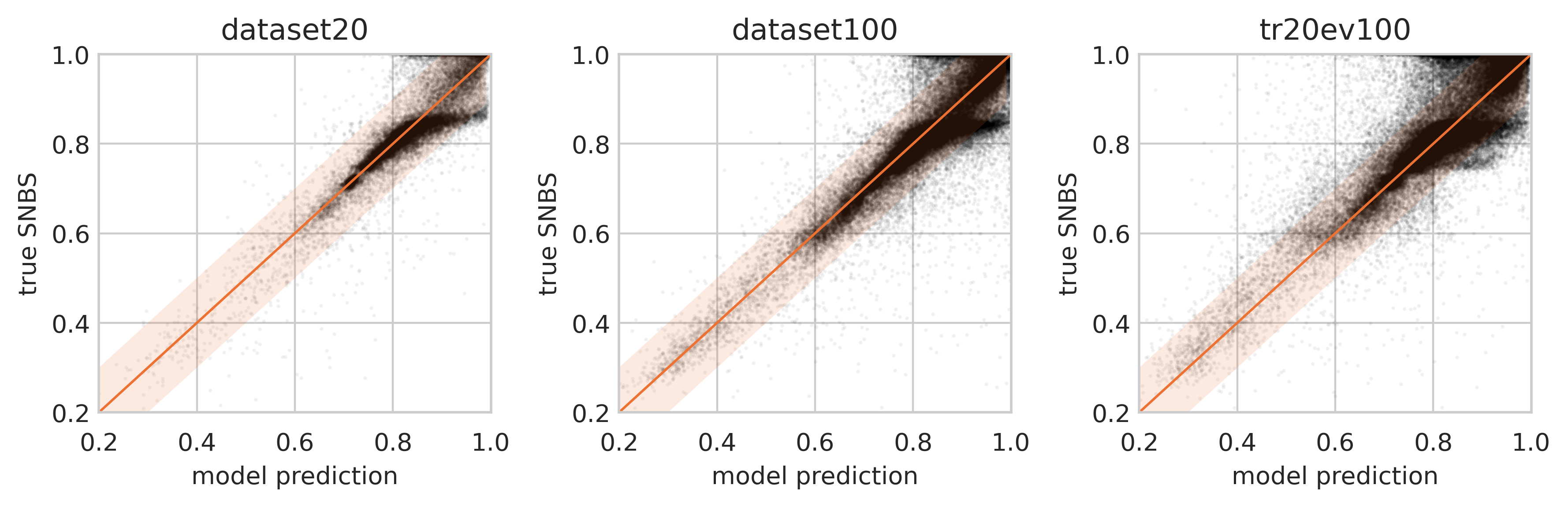}
    \end{subfigure}
    \begin{subfigure}{.49\linewidth}
        \includegraphics[width=\linewidth]{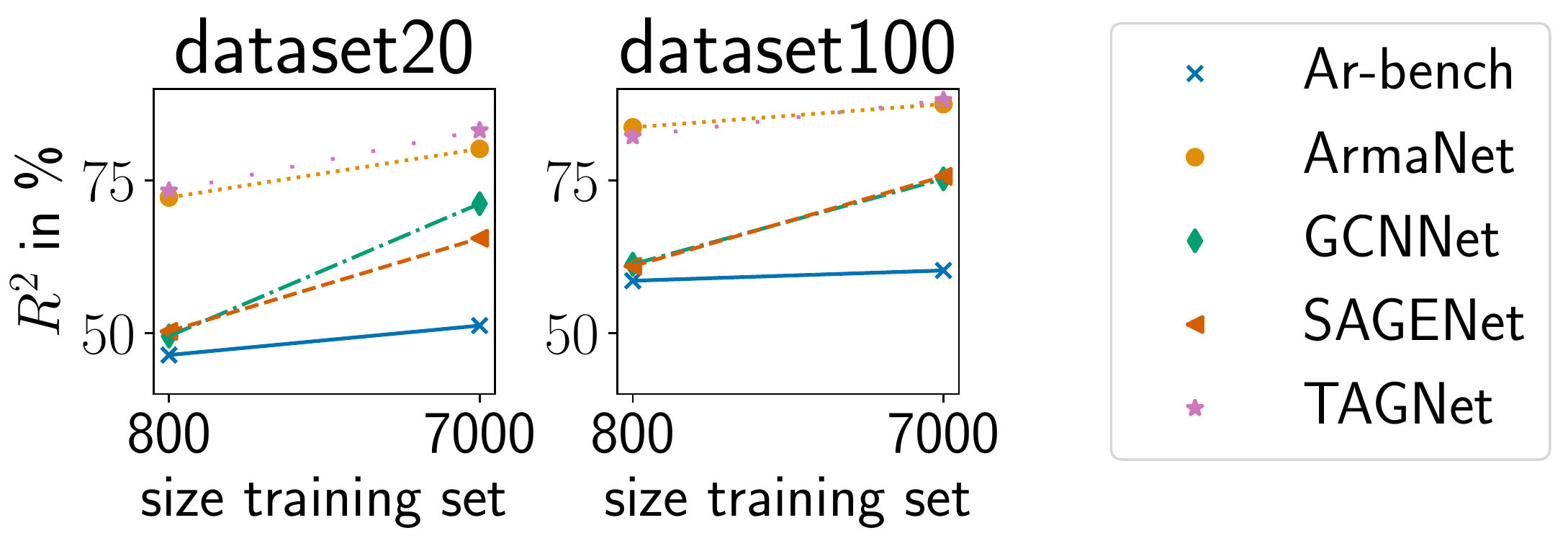}
    \end{subfigure}
        % \includegraphics[width=.6\linewidth]{./pics/prediction_truth_ArmaNet3_dataset_comparison_scatter.pdf}
        % \includegraphics[width=.95\linewidth]{./pics/prediction_truth_ArmaNet3_dataset_comparison_scatter.png}
        % \subfloat{\includegraphics[width=.47\linewidth]{./pics/prediction_truth_ArmaNet3_dataset_comparison_scatter.png}}
        % \caption{t}
        % \subfloat{\includegraphics[width=.47\linewidth]{./pics/more_data_better_performance.pdf}}
    \caption{Left panels: Scatterplot of model prediction and true SNBS values of ArmaNet, the diagonal represents a perfect model ($R^2 = 1$), the banded region indicates prediction errors $\leq 0.1$. Right panels: Influence of size of training data on performance.
    %The three pictures on the left show SNBS over predicted output of the ArmaNet model for dataset20, dataset100 and trained on dataset20, but evaluated on dataset100. . On the right, the two plots compare the performance based on the size of the training set using 800 or 7,000 grids. Training on our larger dataset improves performance on all models. The 800 grids used for the training follow \cite{nauck_predicting_2022} and all models are evaluated on the newly introduced test set.
    \looseness-1}
	\label{fig_result_scatter_more_data}
\end{figure}

% \subsubsection{GNNs can accurately predict SNBS}
\paragraph{GNNs accurately predict SNBS}
GNN models are remarkably successful at predicting the nonlinear target SNBS, both for dataset2: $R^2 > 82 \%$ and for dataset100: $R^2 > 88 \%$,  see the first two columns in \Cref{tb_results_snbs}. Modalities in the data are well captured (left panels in \Cref{fig_result_scatter_more_data}). Both MLPs outperform previous work (Ar-bench), but do not achieve the performance of the newly introduced GNNs.

\begin{table}[h]
    \small
	\centering
	\caption{Results of predicting SNBS represented by $R^2$ score in \%. Each column represents a different setup, e.g. for \textit{tr20ev100} models are trained on dataset20 and evaluated on dataset100.}
	\begin{tabularx}{\linewidth}{XXXXXX}
		\toprule
	   model & tr20ev20 & tr100ev100 & tr20ev100 & tr20evTexas & tr100evTexas \\	  
	   Ar-bench & 51.20 \tiny{$\pm$ 2.762} & 60.24 \tiny{$\pm$ 0.758} & 37.87  \tiny{$\pm$ 2.724} & 40.34 \tiny{$\pm$ 2.833} &  56.86 \tiny{$\pm$ 1.444}\\
		ArmaNet &  80.17 \tiny{$\pm$ 1.226}&  87.50 \tiny{$\pm$ 0.081} & \textbf{68.11} \tiny{$\pm$ 1.933} & 57.09 \tiny{$\pm$ 3.079} & 75.43 \tiny{$\pm$ 0.635}\\
        GCNNet  & 71.18 \tiny{$\pm$ 0.137} & 75.25 \tiny{$\pm$ 0.151} & 58.23  \tiny{$\pm$ 0.059} & -5.29 \tiny{$\pm$ 3.688} & 65.65 \tiny{$\pm$ 0.114} \\
        SAGENet & 65.51 \tiny{$\pm$ 0.253} &  75.66 \tiny{$\pm$ 0.138} &  51.27  \tiny{$\pm$ 0.298} & 32.63 \tiny{$\pm$ 0.515}  & 53.14 \tiny{$\pm$ 2.118}\\
        TAGNet & \textbf{83.19} \tiny{$\pm$ 0.080} & \textbf{88.14} \tiny{$\pm$ 0.081} & 67.00  \tiny{$\pm$ 0.293}&  \textbf{56.05} \tiny{$\pm$ 3.530} & \textbf{82.50} \tiny{$\pm$ 0.438} \\
        \midrule
        linreg & 41.75 & 36.29 & 5.98 & -11.39 & -22.62 \\
        % linreg & 41.75 & 85.71 & 5.98 & 60.52 & -11.39 & 66.02 & 36.29 & 76.58 & -22.62 & 66.60\\
        MLP1 & 58.47  \tiny{$\pm$ 0.149 } & 63.59  \tiny{$\pm$ 0.051}  & 28.49 \tiny{$\pm$ 1.493} & -34.52 \tiny{$\pm$ 17.93} & 19.79  \tiny{$\pm$ 8.659}\\
        MLP2 &  58.20 \tiny{$\pm$ 0.042}&  65.52 \tiny{$\pm$ 0.038} & 19.65 \tiny{$\pm$ 2.109} & 5.81 \tiny{$\pm$10.58 } & 58.46 \tiny{$\pm$ 0.370}\\
		\hline
	 \bottomrule
 	\end{tabularx}
 	\label{tb_results_snbs}
\end{table}

\paragraph{GNNs perform well at out-of-distrubtion tasks}
Using GNNs for SNBS prediction becomes feasible, if they can be trained on relatively simple datasets and still perform well on large, complex grids. Therefore, we define three out-of-distribution tasks: Train on small grids and evaluate on medium-sized grids (tr20ev100), train on small grids and evaluate on the real-sized, synthetic Texan grid (tr20evTexas), and finally train on medium-sized grids and evaluate on the Texan grid (tr100evTexas). The results are shown in \Cref{tb_results_snbs}. For the tr20ev100-task all GNNs generalize well and are able to predict SNBS with up to $R^2 >66 \%$. For tr20evTexas the performance is generally worse, with only ARMANet and TAGNet giving somewhat reliable predictions at up to $R^2 >62 \%$. This loss of performance is probably due to the large differences in grid sizes. Most remarkably for the tr100evTexas tasks, all GNN models perform well (except GCN, see \Cref{sec_appendix_poor_performance_GCN}), with TAGNet reaching an $R^2$ of almost $85\%$. The performance of our models is significantly better when trained on the medium-sized grids, indicating that their topological structure is rich enough to allow for generalization to larger grids. This may be the key for real world applications.
%Training on small grids without loss of generalization and predictive power would be a huge advantage to scale to real power grids. To evaluate the potential of our datasets and GNN models to that end, we apply an out-of distribution task by training the models on dataset20 and evaluating the performance without any further training on dataset100. The third column in \Cref{tb_results_snbs} (tr20ev100) show that all GNNS generalize well and are able to predict SNBS with $R^2$ exceeding 66 \%. We would like to emphasize the significance of that finding. Given sufficient size and complexity in the source dataset, GNNs can robustly predict highly nonlinear stability metrics for grids several times larger than the source. 
%We did not expect grids of size 20 to be large enough to contain enough relevant structures to generalize to larger grids. However, we expect grids of size 100 to generalize well to even larger grids and are encouraged by the out-of-distribution and the Texan power grid (cf. \Cref{sec:Texas})  results. Generalizing from small, numerically solvable grids to large grids is key for real world application. The computational cost of the dynamic simulations grows faster than linearly with the size of the grids, so computational time can be saved when training models on smaller networks or sections of real-sized grids. 
Importantly, the generalization capabilities of the new GNN models are much better than the baselines using MLP or linear regression.\looseness-1
%\paragraph{Predicting SNBS on a Texan power grid model using the previously trained models \label{sec:Texas}}
%Using GNNs for SNBS prediction becomes feasible, if they can be trained on relatively simple datasets and still perform well on large, complex grids. As an example of a large and complex grid, we use a Texan power grid model and evaluate the models previously trained on dataset20 and dataset100. 
%The benchmark models achieve surprisingly high performance with $R^2$ values (above 84 \%) and GCN as a sole exception, see the colums tr20evTexas and tr100evTexas in \Cref{tb_results_snbs}. Details on the poor performance of the GCN is given in \Cref{sec_appendix_poor_performance_GCN}. 
%Hence, the approach of training models on grids which are smaller by more than one order of magnitude is feasible. We want to emphasize that one successful attempt of a real-sized power grid should illustrate the general potential of this approach, but we still consider the hard evidence to be the generalization from 20 to 100. 

%The performance is significantly better for the models trained on dataset100. We hypothesize that the repetition of geometrical structures more prevalent in dataset100 is useful for even larger grids. Grids of size 20 might still be to small to generalize to large grids, but the size 100 might actually be sufficient for many applications.\looseness-1

\paragraph{Training on more data increases the performance of all models} 
Lastly, we evaluate the benefit of more training data. Our experiments (\Cref{fig_result_scatter_more_data} right) show a clear benefit of more training data as compared to \cite{nauck_predicting_2022}, across both grid sizes and all architectures, with differences of up to $\approx20\%$ in $ R^2$. We use the new test set for the comparison of all models. Additional results are given in \Cref{sec_appendix_smaller_dataset}. \looseness-1
\section{Conclusion and Outlook}
This work establishes that GNNs of appropriate size and with enough training data are able to predict probabilistic dynamic stability of models of future power grids with high accuracy. To that end, we provide new datasets 10 times larger than previously published. GNNs trained on the new datasets are able to generalize from small to medium-sized training grids and to real-world sized test grids, promising significant reductions in the simulation time required for grid stability assessment. 
% Future work needs to step up the power grid model detail and further improve the prediction accuracy of GNNs.
The datasets and the code to reproduce the results are published on Zenodo and GitHub, see \Cref{appendix_source_code}. The access enables the community to develop new methods to analyze future renewable power grids. \looseness-1

\subsection*{Acknowledgements}
All authors gratefully acknowledge the European Regional Development Fund (ERDF), the German Federal Ministry of Education and Research, and the Land Brandenburg for supporting this project by providing resources on the high-performance computer system at the Potsdam Institute for Climate Impact Research. Michael Lindner greatly acknowledges support by the Berlin International Graduate School in Model and Simulation (BIMoS) and by his doctoral supervisor Eckehard Schöll. Christian Nauck would like to thank the German Federal Environmental Foundation (DBU) for funding his PhD scholarship and Professor Raisch from Technical University Berlin for supervising his PhD. Special thanks go to Julian Stürmer and his supervisors Mehrnaz Anvari and Anton Plietzsch for their assistance with the Texan power grid model.
% \bibliography{Literature}

\bibliographystyle{unsrt}

\appendix

\section{Appendix}
% Include extra information in the appendix. This section will often be part of the supplemental material. Please see the call on the NeurIPS website for links to additional guides on dataset publication.
This section includes additional information to reproduce the results and also additional results that are not already shown in the main section. We start by providing information on the availability of the data and the used software, followed by details on the evaluation and hyperparameter study and detailed training information for the presented results. Afterwards more results are shown. We also provide details regarding the availability of the datasets and lastly the prediction of SNBS using hand-crafted features, that are considered the baselines in the paper.

\subsection{Availability of the datasets}
\label{appendix_source_code}
The new datasets and full code for the training, evaluation and generation the figures is published on Zenodo (\url{https://zenodo.org/record/7357903}) and GitHub \url{https://github.com/PIK-ICoNe/dynamic_stability_gnn_neurips_climate_workshop.git}. It is licensed under CC-BY 4.0 to enable the community to contribute to this challenge.

\subsection{Software for generating the datasets}
Julia is used for the simulations \cite{bezanson_julia_2017} and the dynamic simulations rely on the package DifferentialEquations.jl \cite{rackauckas_differentialequationsjl_2017}. For simulating more realistic power grids in future work we recommend the additional use of NetworkDynamics.jl \cite{lindner_networkdynamicsjlcomposing_2021} and PowerDynamics.jl \cite{plietzsch_powerdynamicsjl_2021}. 

\subsection{Software for training}
The training is implemented in Pytorch \cite{{paszke_pytorch_2019}}. For the graph handling and graph convolutional layers we rely on the additional library PyTorch Geometric \cite{fey_fast_2019}. We use the SGD-optimizer and as loss function we use the mean squared error \footnote{corresponds to MSELoss in Pytorch}. Furthermore \texttt{ray} \cite{moritz_ray_2018} is used for parallelizing the hyperparameter study. 

\subsection{Coefficient of determination \label{app:r2}}
To evaluate the performance, the coefficient of determination ($R^{2}$-score) is used and computed by $R^2 = 1 - \frac{mse(y,t)}{mse(t_{mean},t)}$, where $mse$ denotes the mean squared error, $y$ the output of the model, $t$ the target value and $t_{mean}$ the mean of all considered targets of the test dataset. $R^2$ captures the mean square error relative to a null model that predicts the mean of the test-dataset for all points. The $R^2$-score is used to measure the portion of explained variance in a dataset. By design, a model that predicts the mean of SNBS per grids has $R^2 = 0$.

\subsection{Hyperparameter optimization}
\label{sec_hyperparameter}
We conduct hyperparameter studies to optimize the model structure regarding number of layers, number of channels and layer-specific parameters using dataset20. The resulting models have the following properties: ArmaNet has 3 layers and 189,048 parameters. GCNNet has 7 layers and 523,020 parameters. SAGENet has 8 layers and 728,869 parameters. TAGNet has 13 layers and 415,320 parameters. Afterwards we optimize learning rate, batch size and scheduler of the best models for dataset20 and dataset100 separately. 

We conduct hyperparameter studies in two steps. First, we optimize model properties such as the number of layers and channels as well as layer-specific parameters e.g. the number of stacks and internal layers in case of ArmaNets. For this optimization we use dataset20 and the SNBS task only. For all models we investigated the influence of different numbers of layers and the numbers of channel between multiple layers. We limit the model size to just above four million parameters, so we did not investigate the full presented space, but limited for example the number of channels when adding more layers.

Afterwards we optimize the learning rate, batch size and scheduler of the best models for dataset20 and dataset100. Hence, our models are not optimized to perform well at the out-of distribution task. As GNN baseline, we use the best model from \cite{nauck_predicting_2022} referred to as Ar-bench, which is a GNN model consisting of 1,050 parameters and based on 2 Arma-layers. The only adjustment to that model is the removal of the fully connected layer after the second Arma-Convolution and before applying the Sigmoid-layer, which improves the training.

\subsection{Details of the training of the benchmark models}
To reproduce the obtained results, more information regarding the training is provided in this section. Detailed information on the training as well as the computation time is shown in \Cref{tb_computation_time_dataset20_dataset100}. In case of dataset20, a scheduler is not applied, in case of dataset100, schedulers are used for Ar-bench (stepLR), GCNNet (ReduceLROnPlateau). The default validation and test set batch size is 150. The validation and test batchsize for Ar-bench and ArmaNet3 is 500 in case of dataset20 and 100 for dataset100. The number of trained epochs differs, because the training is terminated in case of significant overfitting. Furthermore, different batch sizes have significant impact on the simulation time. Most of the training success occurs within the first 100 epochs, afterwards the improvements are relatively small.

\begin{table}[h]
	\centering
	\caption{Properties of training models and regarding the training time, we train 5 seeds in parallel using one nVidia V100.}
    \begin{tabularx}{\linewidth}{XXXXXXXXX}
		\toprule
% 		name & number of epochs & training time & train batch size & learning rate & scheduler \\
    name & \multicolumn{2}{l}{number of epochs} & \multicolumn{2}{l}{training time} & \multicolumn{2}{l}{train batch size} & \multicolumn{2}{l}{learning rate} \\
		dataset& 20 & 100 & 20 (hours) & 100 (days)  & 20 & 100 & 20 & 100\\
		\midrule
		Ar-bench & 1,000 & 800 & 26 & 4 & 200 & 12 & 0.914 & .300 \\ %  None \\
		ArmaNet & 1,500 & 1,000 & 46 & 6 & 228 & 27 & 3.00 & 3.00 \\
		GCNNet & 1,000 & 1000& 29 & 5 & 19 &79 & .307 &.286 \\
		SAGENet & 300 & 1000 & 9 & 5 & 19 & 16 & 1.10 &  1.23 \\
		TAGNet & 400 & 800 & 11 & 4 &  52 & 52 & 0.193 &  .483 \\
% 	 \bottomrule
% 	 f\multicolumn{7}{p{\dimexpr\linewidth-2\tabcolsep-2\arrayrulewidth}}{}
 	\end{tabularx}
	\label{tb_computation_time_dataset20_dataset100}
\end{table}

\subsection{Prediction of SNBS using hand-crafted features}
\label{sec_regresssion_MLP}
Schultz et al. \cite{schultz_detours_2014} predict SNBS based on a regression setup using several hand-crafted features. We use a similar setup to compare this approach to the application of GNNs. Based on the work by Schultz et al. \cite{schultz_detours_2014}, we set up a regression task using the following features: degree, average-neigbor-degree, clustering-coefficient, current-flow-betweenness-centrality and closeness-centrality, as well as the nodal feature $P$. For the regression we use the same train and test split. %The results are given in \Cref{tb_results_snbs}. The performance of the models at the transfer learning tasks is rather poor. In case of tr20ev100, the models has still positive $R^2$-values, but both models have negative $R^2$ values at the prediction of the Texan power grid.

%\paragraph{Application of a Multilayer perceptron}
%In addition, we use the same data as used for the regression to train two Multilayer perceptrons (MLPs). As expected, MLPs outperform the linear regression, but their performance is much lower in comparison to GNNs. MLPs especially do not achieve convincing performances at the out-of-distribution tasks. %In case of the model trained on dataset20, the evalaution results in negative $R^2$, but it works for the model trained on dataset100. This can probably be explained by repeating topological structures in case of grids in dataset100 that also occur in the Texan power grid. However, the performance is still lower in comparison to the GNNs, so we do not expect this approach to work well for a variety of transfer learning tasks.

\paragraph{Training details of MLP}
\label{app_MLP_properties}
The following hyperparmeters are used: MLP1 has one hidden layer with 35 units per hidden layer, resulting in 1,541 parameters and MLP2 has 6 hidden layers and 500 hidden units per layer resulting in 1,507,001 parameters. We conducted hyperparameter studies to optimize the batch sizes and learning rates, see \Cref{tb_hyperparameters_MLP}.

\begin{table}[h]
	\centering
	\caption{Hyperparameters for the MLPs}
	\begin{tabularx}{\linewidth}{XXXX}
		\toprule
	   model & dataset & learning rate  & training batch size\\
	   %& $R^2$ & discr. accu &  $R^2$ & discr. accu & $R^2$ & discr. accu\\
	   %& $R^2$ & 2 &  $R^2$ & 2 & $R^2$ & 2\\
		\hline
		MLP1 & dataset20 & 1.508125539637087 & 1968 \\
		MLP2 & dataset20 & 1.7739583949852091 & 3367 \\
		MLP1 & dataset100 & 1.9519814999342289 & 303 \\
		MLP2 & dataset100 & 0.9978855874564166 & 3768 \\
	 \bottomrule
% 	 \multicolumn{7}{p{\dimexpr\linewidth-2\tabcolsep-2\arrayrulewidth}}{All models are trained on the same 800 grids as in \cite{nauck_predicting_2022}, but evaluated on the newly introduced validation set}
 	\end{tabularx}
 	\label{tb_hyperparameters_MLP}
\end{table}

\subsection{Further Details on power grid generation of the Texan power grid}
\label{app:further_generation}
To take a further step towards real-world applications, we evaluate the performance of our GNN models by analyzing the dynamic stability of a real-sized synthetic model of power grids based on the Texan power grid topology. Real power grid data are not available due to security reasons and calculating an entire SNBS assessment of the fully parameterized synthetic model by \cite{birchfield_activsg2000_nodate} appears not to be feasible due to the computational effort \cite{liemann_probabilistic_2021}. The synthetic grid of Texas is generated using the methods shown in \cite{birchfield_grid_2017}. The Texan power grid model consists of 1,910 nodes after removing 90 nodes that are not relevant for the dispatching. However, that is already a large number and makes the simulations very expensive. the simulation of that grid takes 127,000 CPU hours. Computing less perturbations results in an increased standard error of our estimates of $\pm 0.031$. We use the same modelling approach by arbitrarily modeling nodes as consumers or producers using the $2^{nd}$-order Kuramoto model.

\subsection{Poor performance of GCN when applying to the Texan power grid}
\label{sec_appendix_poor_performance_GCN}
\begin{figure}[h]
    \centering
        \includegraphics[width=.6\linewidth]{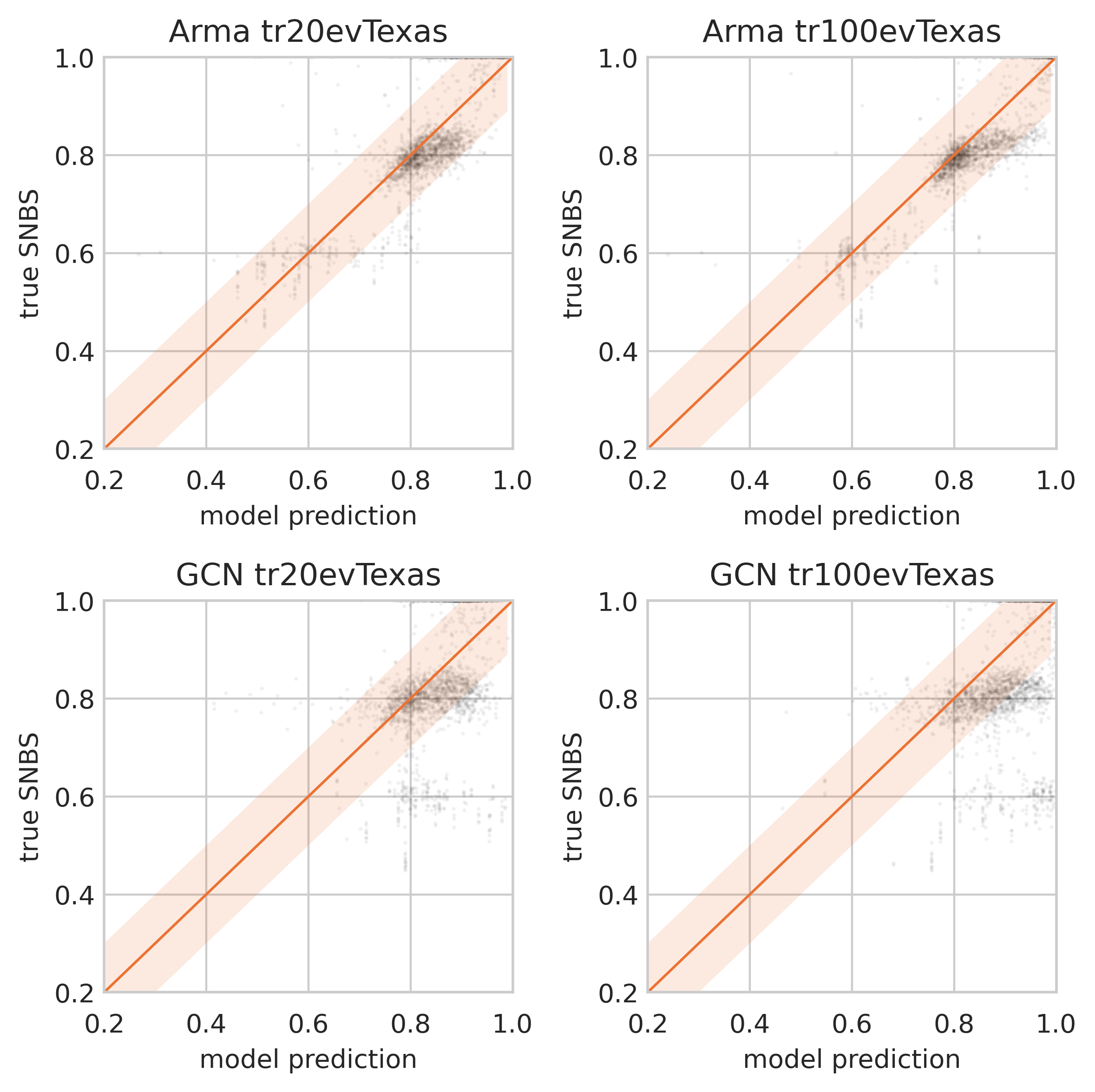}
        % prediction_truth_Texas_ArmaNet_GCNNet_dataset_comparison_scatter_400dpi.png
    \caption{SNBS over predicted output of the Arma and GCN models for the transfer learning task to predict SNBS of the Texan power grid. The diagonal represents a perfect model ($R^2 = 1$), the band indicates the region for accurate predictions based on a tolerance interval of .1 . To account for the small number of nodes, a lower transparency is used in comparison to \Cref{fig_result_scatter_more_data,fig_result_scatter_dataset_comparison_GCNNet7}}
	\label{fig_result_scatter_dataset_comparison_ArmaNet_GCNNet_texas}
\end{figure}

\begin{figure}[h]
    \centering
        \includegraphics[width=.95\linewidth]{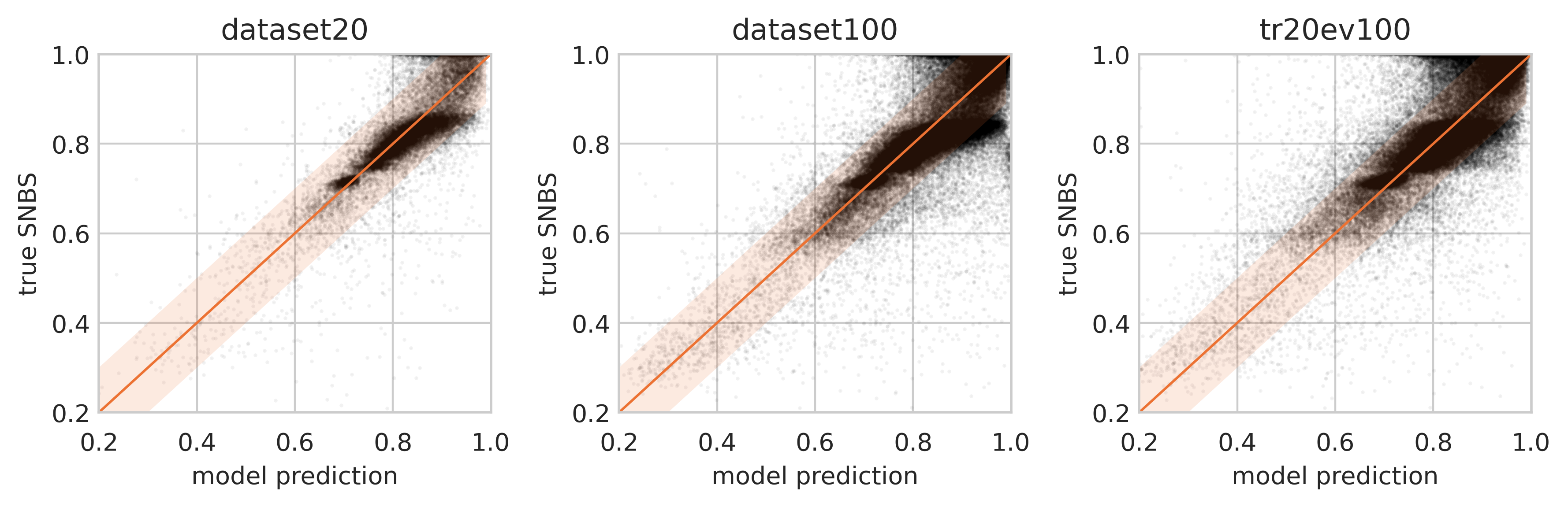}
    \caption{SNBS over predicted output of the GCNNet model for dataset20, dataset100 and trained on dataset20, but evaluated on dataset100. The diagonal represents a perfect model ($R^2 = 1$), the band indicates the region for for accurate predictions based on a threshold of .1}
	\label{fig_result_scatter_dataset_comparison_GCNNet7}
\end{figure}

The GCN model trained on dataset20 is not able to predict the dynamic stability for the Texan power grid. To understand this behaviour, we compare GCNNet and ArmaNet model at the transfer learning task from dataset20 and dataset100 to predict SNBS of the Texan power grid. The scatter plots are shown in \Cref{fig_result_scatter_dataset_comparison_ArmaNet_GCNNet_texas}. We can clearly see that the model is not able to predict lower values of SNBS correctly. The limited output of the GCNNet results in a bad performance in case of the distribution of the Texan power grid that has three modes. As a consequence a model that predicts the mean of the distribution would achieve better performance. Furthermore, we provide the scatter plots of the GCNNet for the three tasks dataset20, dataset100 and tr20ev100 in \Cref{fig_result_scatter_dataset_comparison_GCNNet7}, that can be compared to \Cref{fig_result_scatter_more_data}.
% The GCN model is not able to predict the dynamic stability for the Texan power grid. To understand this behaviour, we compare it to the ArmaNet model at the transfer learning task from dataset20 and dataset100 to predict SNBS of the Texan power grid. The scatter plots are shown in \Cref{fig_result_scatter_dataset_comparison_ArmaNet_GCNNet_texas}. We can clearly see that the model is not able to predict lower values of SNBS correctly. The limited output of the GCNNet results in a bad performance in case of the distribution of the Texan power grid that has three modes. As a consequence a model that predicts the mean of the distribution would achieve better performance. Furthermore, we provide the scatter plots of the GCNNet for the three tasks dataset20, dataset100 and tr20ev100 in \Cref{fig_result_scatter_dataset_comparison_GCNNet7}, that can be compared to \Cref{fig_result_scatter_more_data}.

\subsection{Detailed results of training on a smaller dataset}
\label{sec_appendix_smaller_dataset}
To investigate the influence of available training data and to connect with previous work, we train all models on only 800 grids, from \cite{nauck_predicting_2022}. The results are shown in \Cref{tb_ResultsR2score_small_data}. 

% \begin{table}[h]
% submitted
% 	\centering
% 	\caption{Performance after training on smaller training set. All models are trained on the same 800 grids as in \cite{nauck_predicting_2022}, but evaluated on the newly introduced test set. The results are represented by $R^2$ score in \%.}
% 	\begin{tabularx}{\linewidth}{XXXX}
% 		\toprule
% 	   model & dataset20 & dataset100 & tr20ev100\\
% 	   %& $R^2$ & discr. accu &  $R^2$ & discr. accu & $R^2$ & discr. accu\\
% 	   %& $R^2$ & 2 &  $R^2$ & 2 & $R^2$ & 2\\
% 		\hline
% 		Ar-bench & 46.54 \tiny{$\pm$ 2.378} & 59.73 \tiny{$\pm$ 0.886} & 31.75 \tiny{$\pm$ 1.204} \\
% 		ArmaNet & 70.35 \tiny{$\pm$ 1.226}& \textbf{83.92} \tiny{$\pm$ 0.263}& 55.84  \tiny{$\pm$ 0.598}   \\
%         GCNNet & 49.59 \tiny{$\pm$ 0.513}  & 61.18 \tiny{$\pm$ 1.663}& 36.08 \tiny{$\pm$ 0.625}   \\
%         SAGENet & 50.15 \tiny{$\pm$ 0.255} & 60.98 \tiny{$\pm$ 0.279} & 39.89  \tiny{$\pm$ 0.089} \\
%         TAGNet   & \textbf{74.77} \tiny{$\pm$ 0.370} &  82.21 \tiny{$\pm$ 0.017}  &  \textbf{60.31} \tiny{$\pm$ 0.732}\\
% 	 \bottomrule
% % 	 \multicolumn{7}{p{\dimexpr\linewidth-2\tabcolsep-2\arrayrulewidth}}{All models are trained on the same 800 grids as in \cite{nauck_predicting_2022}, but evaluated on the newly introduced validation set}
%  	\end{tabularx}
%  	\label{tb_ResultsR2score_small_data}
% \end{table}

\begin{table}[h]
	\centering
	\caption{Performance after training on smaller training set. All models are trained on the same 800 grids as in \cite{nauck_predicting_2022}, but evaluated on the newly introduced test set. The results are represented by $R^2$ score in \%.}
	\begin{tabularx}{\linewidth}{XXXX}
		\toprule
	   model & dataset20 & dataset100 & tr20ev100\\
	   %& $R^2$ & discr. accu &  $R^2$ & discr. accu & $R^2$ & discr. accu\\
	   %& $R^2$ & 2 &  $R^2$ & 2 & $R^2$ & 2\\
		\hline
		Ar-bench & 46.38 \tiny{$\pm$ 2.355} & 58.55 \tiny{$\pm$ 1.918} & 31.75 \tiny{$\pm$ 1.204} \\
		ArmaNet & 72.20 \tiny{$\pm$ 1.168}& \textbf{83.70} \tiny{$\pm$ 0.220}& 54.12  \tiny{$\pm$ 3.187}   \\
        GCNNet & 49.48 \tiny{$\pm$ 0.247}  & 61.26 \tiny{$\pm$ 1.158}& 39.59 \tiny{$\pm$ 0.285}   \\
        SAGENet & 50.26 \tiny{$\pm$ 0.450} & 60.94 \tiny{$\pm$ 0.167} & 38.93  \tiny{$\pm$ 0.902} \\
        TAGNet   & \textbf{73.30} \tiny{$\pm$ 0.304} &  82.21 \tiny{$\pm$ 0.143}  &  \textbf{61.47} \tiny{$\pm$ 0.462}\\
	 \bottomrule
% 	 \multicolumn{7}{p{\dimexpr\linewidth-2\tabcolsep-2\arrayrulewidth}}{All models are trained on the same 800 grids as in \cite{nauck_predicting_2022}, but evaluated on the newly introduced validation set}
 	\end{tabularx}
 	\label{tb_ResultsR2score_small_data}
\end{table}

\end{document}